\documentclass[letterpaper, 10 pt, conference]{ieeeconf}  % Comment this line out if you need a4paper

\IEEEoverridecommandlockouts                              % This command is only needed if 
\usepackage[T1]{fontenc}

\usepackage{amssymb}
\usepackage{amsmath}
\usepackage{color, soul}
\usepackage{booktabs}
\usepackage{colortbl}
\usepackage{tcolorbox}
\usepackage{color,xcolor}
\usepackage{microtype}
\usepackage{inconsolata}
\usepackage{fancyvrb}
\usepackage{multirow}
\usepackage{colortbl}
\usepackage{lipsum}
\usepackage{pgfplots}
\usepackage{siunitx}
\usepackage{hyperref}
\usepackage{subcaption}
\usepackage{algorithm}
\usepackage{algpseudocode}
\hypersetup{
    colorlinks=true,
    linkcolor=black,
    filecolor=black,
    urlcolor=black,
    citecolor=black,
}
\usepackage{float} % 这个包允许我们使用 [b!] 选项
\usepackage{newunicodechar}
\newunicodechar{～}{\textasciitilde}

\title{\LARGE \bf TimeLDM: Latent Diffusion Model for Unconditional Time Series Generation}

\author{Jian Qian$^{1}$, Bingyu Xie$^{2}$, Biao Wan$^{1}$, Minhao Li$^{1}$, Miao Sun$^{3}$ and Patrick Yin Chiang$^{1}$  % <-this % stops a space
\thanks{$^{1}$ Fudan University,  \texttt{\footnotesize \{jqian20, pchiang\}@fudan.edu.cn.}}
\thanks{$^{2}$ Carnegie Mellon University, \texttt{\footnotesize vxie@andrew.cmu.edu.}}
\thanks{$^{3}$ Nanyang Technological University, \texttt{\footnotesize miao.sun@ntu.edu.sg.}}
}

\begin{document}

\maketitle
\thispagestyle{empty}
\pagestyle{empty}

%%%%%%%%%%%%%%%%%%%%%%%%%%%%%%%%%%%%%%%%%%%%%%%%%%%%%%%%%%%%%%%%%%%%%%%%%%%%%%%%
\begin{abstract}

Time series generation is a crucial research topic in the area of decision-making systems, which can be particularly important in domains like autonomous driving, healthcare, and, notably, robotics. 
Recent approaches focus on learning in the data space to model time series information. However, the data space often contains limited observations and noisy features.
In this paper, we propose TimeLDM, a novel latent diffusion model for high-quality time series generation.
TimeLDM is composed of a variational autoencoder that encodes time series into an informative and smoothed latent content and a latent diffusion model operating in the latent space to generate latent information.
We evaluate the ability of our method to generate synthetic time series with simulated and real-world datasets and benchmark the performance against existing state-of-the-art methods.
Qualitatively and quantitatively, we find that the proposed TimeLDM persistently delivers high-quality generated time series.
For example, TimeLDM achieves new state-of-the-art results on the simulated benchmarks and an average improvement of 55\% in Discriminative score with all benchmarks.
Further studies demonstrate that our method yields more robust outcomes across various lengths of time series data generation. Especially, for the Context-FID score and Discriminative score, TimeLDM realizes significant improvements of 80\% and 50\%, respectively. 
The code will be released after publication.
% Currently, latent diffusion models are ascending to the forefront of generative modeling for many important data representations. 
% Being the most pivotal in the computer vision domain, latent diffusion models have also recently attracted interest in other communities, including NLP, Speech, and Geometric Space.
% Sores from Context-FID and Discriminative indicate that TimeLDM consistently and significantly outperforms current state-of-the-art benchmarks with an average improvement of 3.4$\times$ and 3.8$\times$, respectively.
% Further studies demonstrate that our method presents better performance on different lengths of time series data generation.
% To the best of our knowledge, this is the first study to explore the potential of the latent diffusion model for unconditional time series generation and establish a new baseline for synthetic time series.

\end{abstract}

%%%%%%%%%%%%%%%%%%%%%%%%%%%%%%%%%%%%%%%%%%%%%%%%%%%%%%%%%%%%%%%%%%%%%%%%%%%%%%%%
\section{INTRODUCTION}
Time series generation holds a pivotal role across numerous applications, such as robotics~\cite{ichiwara2023multimodal,hu2022causal}, autonomous driving~\cite{deng2021deep,huang2022survey}, and healthcare~\cite{afzal2024rest, choi2024feasibility}.
Additionally, generating time series can be a valuable approach to solving the complex challenges associated with data privacy concerns. 
It enables agents to learn a wealth of information without containing any actual sensitive data, providing a safer framework for model training and development. 

Numerous studies have used various architectures of deep neural networks for synthetic realistic time series data,
including Variational Autoencoder (VAE) based methods~\cite{desai2021timevae,fortuin2020gp}, Generative Adversarial Network (GAN) based methods~\cite{yoon2019time,pei2021towards,jeha2021psa,xu2020cot}, and Diffusion-based methods~\cite{coletta2024constrained,yuan2024diffusion}.
% and more~\cite{jarrett2021time,fons2022hypertime}.
Typically, Diffusion-based methods have gained plenty of attention from researchers. 
For instance,
% % Cot-GAN~\cite{xu2020cot} incorporates a specialized loss function based on a regularized Sinkhorn distance, which originates from the principles of causal optimal transport theory.
% % TimeGAN~\cite{yoon2019time} merges the adaptability of the unsupervised GAN framework with the control afforded by supervised training in autoregressive models. However, despite the potential application, the results of the generation are still unsatisfactory compared to simulated and realistic time series data. 
% Most recently, diffusion models (DMs)~\cite{ho2020denoising, song2020score} have shown impressive performance on image generation~\cite{saharia2022palette,dhariwal2021diffusion} and beyond~\cite{croitoru2023diffusion,yang2023diffusion}. 
% DMs construct a diffusion process that gradually degrades the training data structure, and learned neural networks to reverse this contaminated data by progressive denoising. 
% To further improve the synthesis quality of time series data, several studies have also applied DMs for time series synthesis, such as 
DiffTime~\cite{coletta2024constrained} adopts future mix-up and autoregressive initialization as a condition to generate time information. 
Diffusion-TS~\cite{yuan2024diffusion} combines the interpretability component, such as trend and multiple seasonality, to model time series using denoising diffusion models.
Those methods have emerged as a superior learning architecture in generative modeling to others.
However, existing approaches often apply learning models directly in the data space, which typically consists of \textit{limited information and noisy features}. 
Therefore, we are interested in searching for a more flexible framework for time series modeling.  
\begin{figure}[t]
\centering
\resizebox{0.8\columnwidth}{!}{
\includegraphics{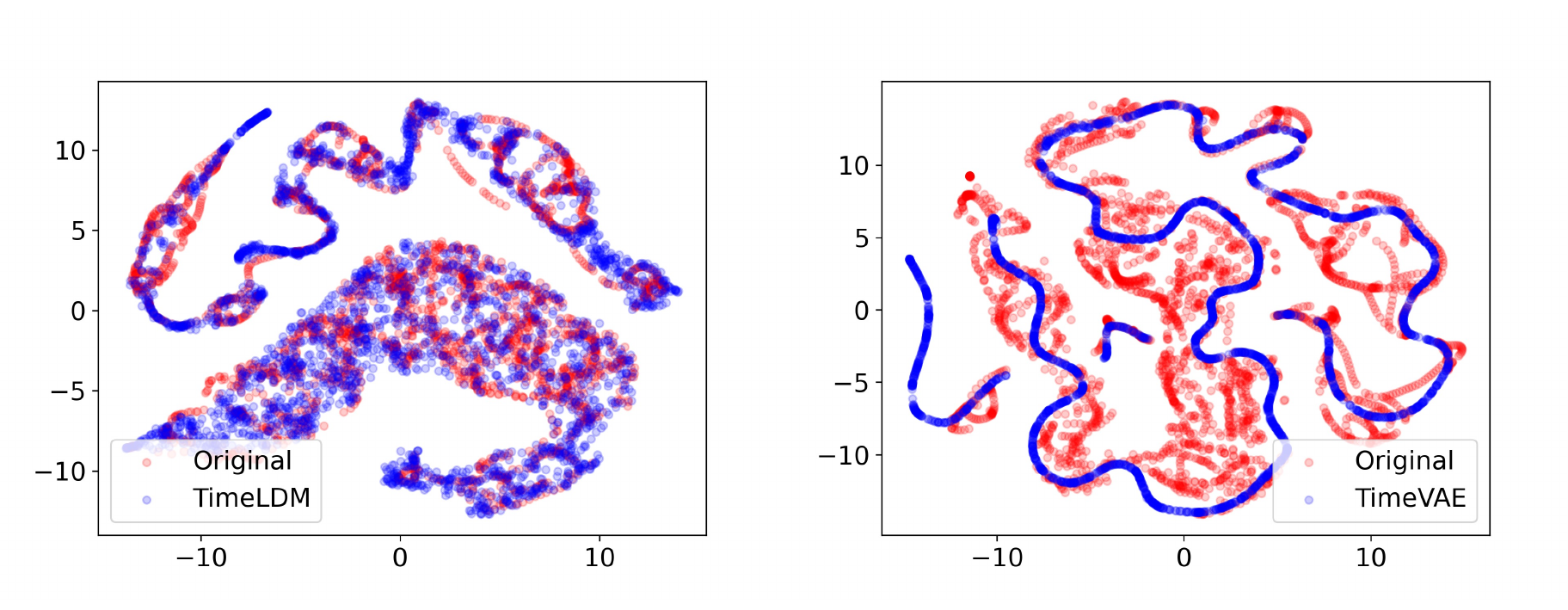}}
\caption{t-SNE visualization on the stocks dataset, TimeLDM shows better overlap between the generated data and original data than TimeVAE.} 
\label{fig:kernel1}
\end{figure}
% where a greater overlap of blue and red dots shows a better distributional similarity between the generated data and original data.
\begin{figure*}[h]
\centering
\resizebox{0.95\textwidth}{!}{%
\includegraphics{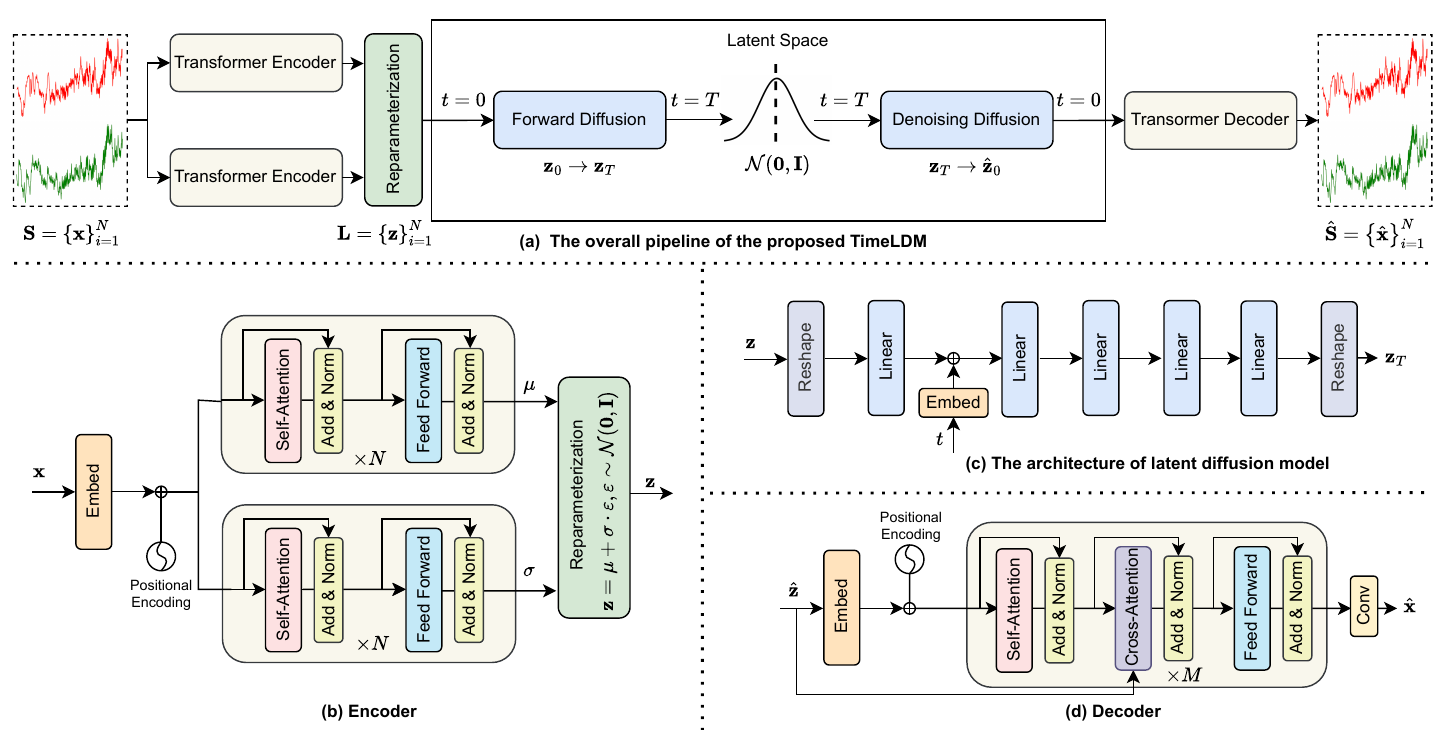}}
\caption{Structure of our proposed TimeLDM. (a) shows the components of TimeLDM, consisting of the transformer encoder, reparameterization, diffusion process, reverse process, and transformer decoder. (b) shows the details of the transformer encoder and reparameterization. (c) shows the architecture of the latent diffusion model. (d) shows the details of the transformer decoder.}
% % \vspace{10pt} 
\label{fig:timeLDM}
\end{figure*}

Latent space generation, an efficient alternative model in generative architectures, adopts a pre-trained autoencoder to transfer the generation tasks from the input space to a greater flexible latent domain.
In this paper, inspired by the success of the diffusion model on latent space~\cite{rombach2022high,li2022improving}, we propose an efficiently synthesized time series method to overcome the above limitations by adopting a \textit{smoother and informative latent presentation}, named \textbf{TimeLDM} (\textbf{Time} \textbf{L}atent \textbf{D}iffusion \textbf{M}odel). 
% Specifically, we adopt the LDMs within a framework comprising a variational autoencoding model that projects time data into a latent space, and a denoising MLP that performs a reverse diffusion process in that latent space. 
As shown in Figure~\ref{fig:timeLDM} (a),
We first transform the raw time series data into an embedding space and train the encoder and decoder network for the VAE. 
The well-studied VAE converts the time series data into the latent space.
After that, we apply the latent information as the target of the latent diffusion model (LDM), which is designed with a denoising MLP. 
During inference, we generate the latent vectors from the LDM and then apply the VAE decoder to synthesize the time series. 

We validate the performance of our proposed approach for different benchmarks, including simulated and real-world time series datasets. 
Qualitatively and quantitatively, we find that
the proposed TimeLDM persistently delivers high-quality generated
time series (see Figure～\ref{fig:kernel4}).
In Table~\ref{tab:unconditional-TS1} and~\ref{tab:unconditional-TS2}, 
the Discriminative scores of TimeLDM consistently outperform current state-of-the-art benchmarks.
Furthermore, Table~\ref{tab:lunconditional-TS1} demonstrates that TimeLDM
presents better performance on different lengths of time series data
generation. The main contributions of this paper are summarized as follows:
\begin{itemize}
\item  We propose {TimeLDM}, a latent diffusion-based method that leverages the high-fidelity image synthesis ability into unconditional time series generation. To the best of our knowledge, this is the first work to explore the potential of LDM for unconditional time series generation. 
\item  We evaluate the ability of our method to synthetic time series with simulated and real-world datasets. 
% benchmark the performance against existing state-of-the-art methods. 
Empirically, TimeLDM shows better performance than existing generation methods both qualitatively and quantitatively. The ablation study presents the proposed loss function of VAE, which plays a crucial role in improving the capability of our method.
\item Furthermore, we evaluate TimeLDM with different lengths of time series data, which presents better performance on the proposed benchmark datasets compared with current state-of-the-art methods.
\end{itemize}

%%%%%%%%%%%%%%%%%%%%%%%
\section{RELATED WORKS}
%%%%%%%%%%%%%%%%%%%%%%%
\noindent\textbf{Time Series Generation.}
Deep generative models have demonstrated their ability to generate high-quality samples across a wide array of fields, where generating time series stands as a particularly challenging endeavor within limited information and noisy features.
Early methods based on GANs~\cite{goodfellow2014generative} have been extensively investigated for time series generation. 
% The GAN comprises two neural networks that are trained concurrently: a discriminator trained to differentiate between real samples from the training dataset and those that are generated,
% and a generator that is trained to produce samples that “fool” the discriminator.  
For example, 
TimeGAN~\cite{yoon2019time} applies an embedding function and supervised loss to the original GAN for capturing the temporal dynamics of data throughout time. 
% RTSGAN~\cite{pei2021towards} and PSA-GAN~\cite{jeha2021psa} generate long univariate time series samples of high quality with self-attention mechanism.
Cot-GAN~\cite{xu2020cot} incorporates a specialized loss function based on a regularized Sinkhorn distance, which originates from the principles of causal optimal transport theory.
While VAEs~\cite{kingma2013auto} also drew the attention of researchers.
For instance,
% Fourier Flows~\cite{alaa2020generative} is proposed as a method based on normalizing flows followed by a chain of spectral filters leading to an exact likelihood optimization. 
TimeVAE~\cite{desai2021timevae} implements an interpretable temporal structure and achieves reasonable results on time series synthesis. Recent research~\cite{ coletta2024constrained,yuan2024diffusion} has been exploring the use of diffusion models~\cite{ho2020denoising} to generate time series, developing on the successes of forward and reverse processing in other areas such as images~\cite{gu2022vector}, video~\cite{luo2023videofusion}, text~\cite{yu2022latent}, and audio~\cite{kong2020diffwave}.  
Among them, 
% TimeGrad~\cite{rasul2021autoregressive} is a conditional diffusion model that predicts in an autoregressive manner, with the denoising process guided by the hidden state of a recurrent neural network. 
% CSDI~\cite{tashiro2021csdi} uses self-supervised masking to guide the denoising process like image inpainting.
% SSSD~\cite{alcaraz2022diffusion} based on the structured state space model. 
DiffTime~\cite{coletta2024constrained} approximated the diffusion function based on CSDI~\cite{tashiro2021csdi} where they remove the side information provided as embedding. 
% They also introduced guided DiffTime to handle new constraints such as trend or fixed-value, without re-training. 
Diffusion-TS~\cite{yuan2024diffusion} combines the interpretability component, such as trend and multiple seasonality, to model time series using denoising diffusion models.

%%%%%%%%%%%%%%%%%%%%%%%
\noindent\textbf{Generative Modeling in the Latent Space.}
Although generative models in the data space have achieved significant success, the latest emerging LDMs~\cite{rombach2022high,li2022improving} have demonstrated several advantages, including more compact and disentangled representations, robustness to noise, and greater flexibility in controlling generated styles. 
% To enable faster training and inference on limited computational resources.
% Due to the compaction and smoothness of the latent space, LDMs are preferred for their computational efficiency. 
LDMs have achieved great success in image generation as they exhibit better scaling properties and expressivity than the vanilla diffusion models in the data space.
The success of the LDM in image generation has also inspired their applications in video~\cite{blattmann2023align}, audio~\cite{liu2023audioldm}, tabular~\cite{zhang2023mixed}, and text~\cite{lovelace2024latent} domains. 
In this paper, we explore the application of the LDM for unconditional time series generation tasks.

\renewcommand{\algorithmicrequire}{\textbf{Input:}} 
\renewcommand{\algorithmicensure}{\textbf{Output:}} 
% {\footnotesize
\begin{algorithm}[b!]
\caption{Training Algorithm of TimeLDM}
\begin{algorithmic}
\small % 这里可以用 \small, \footnotesize, 或者 \scriptsize 来调整大小
\Require Time series data $\mathbf{S}=\left\{\mathbf{x}\right\}_{i=1}^N$
\Ensure Encoder $\mathcal{E}_\phi$, Decoder $\mathcal{D}_{\xi}$,  Denoising Network $\boldsymbol{\epsilon}_\theta$
\Function{Train AutoEncoder}{}
    \State Initialize $\mathcal{E}_\phi, \mathcal{D}_{\xi}$
    \While{$\phi, \xi$ have not converged}
        \State Sample ${\mathbf{x}} \in {{\mathbf{S}}}$
        \State Get the embedding pattern
        \State Get the positional pattern
        \State $\mu,\sigma  \leftarrow \mathcal{E}_\phi(\mathbf{x})$
        \State $\varepsilon \sim \mathcal{N}(\mathbf{0}, \mathbf{I})$
        \State $\text{Reparameterization}: \mathbf{z}={\mu}+\varepsilon \cdot \sigma$
        \State $\hat{\mathbf{x}}\leftarrow \mathcal{D}_{\xi}\left(\mathbf{z}\right)$
        \State $\mathcal{L}=\mathcal{L}_{\text {recon}}(\mathbf{x}, \hat{\mathbf{x}})+\beta \mathcal{L}_{\text {KL}}({\mu}, {\sigma})$
        \State $\phi, \xi \leftarrow \operatorname{optimizer}\left(\mathcal{L} ; \phi, \xi\right)$
         \If {$\ell_{\text{recon}}$ fails to decrease for $S$ steps}
                \State $\beta \leftarrow \lambda \beta$
        \EndIf
    \EndWhile
    \State \Return $\mathcal{E}_\phi, \mathcal{D}_{\xi}$
\EndFunction
\Function{Train Latent Diffusion}{}
    \State Initialize $\epsilon_\theta$
    \While{$\theta$ have not converged}
        \State $\mathbf{z} \sim q_\phi\left(\mathbf{z} \mid \mathbf{x} \right)$ %\Comment{As lines}
        \State $t \sim \mathbf{U}(0, T)$
        \State $\varepsilon \sim \mathcal{N}\left(\mathbf{0}, \sigma^2 \mathbf{I}\right)$
        \State $\mathbf{z}_t=\mathbf{z}_0+\varepsilon$
        \State $\ell(\theta)=\left\|\epsilon_\theta\left(\mathbf{z}_t, t\right)-\varepsilon\right\|_2^2$
        \State $\theta \leftarrow \operatorname{optimizer}\left(\mathcal{L}_{L D M} ; \theta\right)$
    \EndWhile
    \State \Return $\epsilon_\theta$
\EndFunction
\State $\mathcal{E}_\phi, \mathcal{D}_{\xi} \leftarrow$ \Call{Train AutoEncoder}{}
\State Fix parameters $\phi$ and $\xi$
\State $\epsilon_\theta \leftarrow$ \Call{Train Latent Diffusion}{}
\State \Return $\mathcal{E}_\phi, \mathcal{D}_{\xi}, \epsilon_\theta$
\end{algorithmic}
\label{ag:training}
\end{algorithm}
% }
%%%%%%%%%%%%%%%%%%%%%%%%%%%%%%%%%%%%%%%%%%%%%%%%%%%%%%%%%%%%%%%%%%%%%%%%
\section{METHOD}

TimeLDM, as shown in Figure~\ref{fig:timeLDM}, consists of an encoder-decoder module for VAE, a reparameterization trick for latent information sampling, and the LDM. In this section,
we formulate the time series generation task first.
Then, we introduce the details of VAE and LDM from network architecture to mathematical formulation.
Finally, we summarize the training and sampling procedures.

%%%%%%%%%%%%%%%%%%%%%%%
\subsection{Problem Statement}
\label{sec:ps}
Let $\mathbf{x}_{1: \tau}=\left(x_1, \ldots, x_\tau\right) \in \mathbb{R}^{\tau \times d}$ be the original  time series, where $\tau$ denotes time steps, $d$ is the dimension of observed signals. 
Given the time series dataset $\mathbf{S}=\left\{\mathbf{x}\right\}_{i=1}^N$, 
the aim of TimeLDM is to learn parameterized generative model $p_\theta(\mathbf{S})$, which can accurately synthesize diverse and realistic time series data $\hat{\mathbf{x}} \in \hat{{\mathbf{S}}}$ without condition.

%%%%%%%%%%%%%%%%%%%%%%%
\subsection{Time Series Autoencoding}
\label{sec:vae}
To overcome the weakness of data domain generation, we are focusing on presenting the time series signals 
$\mathbf{S}=\left\{\mathbf{x}\right\}_{i=1}^N$ into an informative and smoothed latent space.
The latent representation is $\mathbf{L}=\left\{\mathbf{z}\right\}_{i=1}^N$, where $\mathbf{z}_{1: \tau}=\left(z_1, \ldots, z_\tau\right) \in \mathbb{R}^{\tau \times m}$ denotes the latent feature and $m$ is the  dimension of representation. The framework of VAE is shown in Figure~\ref{fig:timeLDM} (b) and (d). As we can see, it designs with encoder and decoder module, VAE $=(\mathcal{E}_\phi(\mathbf{x}), \mathcal{D}_{\xi}(\mathbf{z}))$, where the encoder $\mathcal{E}_\phi$ learn the latent variable $\mathbf{z} = \mathcal{E}_\phi(\mathbf{x})$ and 
 the decoder $\mathcal{D}_{\xi}$ decode latent feature $\mathbf{z}$ back to data domain $\hat{\mathbf{x}}=\mathcal{D}_{\xi}({\mathbf{z}})$. 
% Unlike the vanilla VAE apply the ELBO loss function to train, 
Here we adopt $\beta$-VAE~\cite{higgins2017beta}, the coefficient $\beta$ adaptively balances the reconstruction loss and KL-divergence loss for effective training. 

% \vspace{10pt}
\noindent\textbf{\noindent\textbf{VAE’s Encoder.}} As shown in Figure~\ref{fig:timeLDM} (b), the VAE's Encoder $\mathcal{E}_\phi$ first apply a convolutional neural network to learn an embedding pattern $\mathbf{e} = emb(\mathbf{x}) \in \mathbb{R}^{\tau \times m} $ from the temporal structures $\mathbf{x}_{1: \tau} \in \mathbb{R}^{\tau \times d}$, then a learnable positional encoding $pe\in \mathbb{R}^{\tau \times m} $ equip to the embedding feature for adaptively learning the time series positional information. After that, we train two transformer encoders to learn the mean $\mu\in \mathbb{R}^{\tau \times m}$ and log variance $\sigma\in \mathbb{R}^{\tau \times m}$ from the positional encoding feature $\mathbf{e}^{pe}_{1: \tau} = \mathbf{e}_{1: \tau} + pe$, respectively. 
Next, we obtain the latent variables $\mathbf{z}_{1: \tau}$ from the reparameterization trick  Function~\ref{eq:ldm}. 

\begin{equation}
\mathbf{z}=\mu+\sigma \cdot \varepsilon,  \varepsilon \sim \mathcal{N}(\mathbf{0}, \mathbf{I})
\label{eq:ldm}
\end{equation}

% \vspace{10pt} 
\noindent\textbf{\noindent\textbf{VAE’s Decoder.}} As shown in Figure~\ref{fig:timeLDM} (d), 
the aim of VAE’s Decoder $\mathcal{D}_{\xi}$ is to minimize the reconstruction error by generating outputs that closely resemble the original time series information. The input to the Decoder  $\mathcal{D}_{\xi}$ consists of latent variables sampled from a typically Gaussian distribution, which is derived from the Encoder using the reparameterization trick.
The architect of the VAE decoder incorporates both self-attention and cross-attention mechanisms. The input also respects the learning embedding and positional encoding process. Finally, it generates the realistic samples of time series data $\hat{\mathbf{x}}=\mathcal{D}_{\xi}({\mathbf{z}})$.

% \vspace{10pt} 
\noindent\textbf{\noindent\textbf{Training Loss.}}
The training objective of the VAE consists of the reconstruction loss $\mathcal{L}_\text{recon}$ and the KL divergence $\mathcal{L}_\text{KL}$.
The reconstruction loss is composed of the $\mathcal{L}_1$ norm, $\mathcal{L}_2$ norm in the data domain, and the Fast Fourier Transformation (FFT)~\cite{Nussbaumer1982} loss term $\|\mathcal{F F} \mathcal{T}(\mathbf{x}), \mathcal{F F} \mathcal{T}(\hat{\mathbf{x}})\|$  in the frequency domain~\cite{bracewell1986fourier}, which is
% between the input data and the reconstructed signal. 
inspired by HyperTime~\cite{fons2022hypertime} for accurate time series reconstruction. $\lambda_1$, $\lambda_2$, and $\lambda_3$ are weights to balance three losses.
\begin{equation}
\centering % 确保公式居中
\resizebox{0.9\columnwidth}{!}{$
\mathcal{L}_{\text {recon}}=\lambda_1\|\mathbf{x}-\hat{\mathbf{x}}\|_2^2 + \lambda_2\|\mathbf{x}-\hat{\mathbf{x}}\| + \lambda_3\|\mathcal{FFT}\left(\mathbf{x}\right), \mathcal{FFT}\left(\hat{\mathbf{x}}\right)\| 
$}
\label{eq:reconstruction loss}
\end{equation}
\noindent  KL divergence loss regularizes the mean and log variance of the latent space. As shown Equation~\ref{eq:KLD}, the $q_\phi(\mathbf{z} \mid \mathbf{x})$ is probabilistic output from the encoder $\mathcal{E}_\phi$ that represents the approximate posterior of latent variable $\mathbf{z}$ given the input $\mathbf{x}$; $\mathcal{N}(\mathbf{z} ; \mu, \sigma)$ is the prior on $\mathbf{z}$ . The $\beta$ is adaptively tuned during training, where $\beta=\lambda \beta, \lambda<1$. If the  $\mathcal{L}_{\text {recon}}$ fails to decrease with defined steps, the $\beta$ will decrease to encourage the model to pay more attention to the reconstruction term.
\begin{equation}
\mathcal{L}_\text{KL}= \beta \mathrm{KL}\left(q_{\boldsymbol{\phi}}(\boldsymbol{\mathbf{z}} \mid \mathbf{x}) \| \mathcal{N}(\mathbf{z} ; \mu, {\sigma})\right)
\label{eq:KLD}
\end{equation}
\noindent Finally, the overall training objective of the VAE is as below. For the adaptive $\beta$, we set $\beta_{\max }=10^{-2}$, $\beta_{\min }=10^{-5}$, and $\lambda=0.7$, where the $\beta_{\max }$ is initial setting, and $\beta_{\min }$ is the minimum number of the adaptive $\beta$. 
\begin{equation}
\mathcal{L}=\mathcal{L}_{\text {recon}} + \mathcal{L}_\text{KL}
\label{eq:KLD1}
\end{equation}

\renewcommand{\algorithmicrequire}{\textbf{Input:}} 
\renewcommand{\algorithmicensure}{\textbf{Output:}} 
\begin{algorithm}[tbp]
\caption{Sampling Algorithm of TimeLDM}
\begin{algorithmic}
\small
\Require Decoder network $\mathcal{D}_{\xi}$, denoising network $\boldsymbol{\epsilon}_\theta$
\Ensure $\hat{\mathbf{x}} \in \hat{{\mathbf{S}}}$
\State Sample $\mathbf{z}_T \sim \mathcal{N}\left(\mathbf{0}, \sigma^2(T) \mathbf{I}\right), t_{\max }=T$
\For{$i = \text{max}, \dots, 1$}
    \State $\nabla_{\mathbf{z}_{t_i}} \log p(\mathbf{z}_{t_i}) = -\mathbf{\epsilon}_\theta(\mathbf{z}_{t_i}, t_i) / \sigma(t_i)$
    \State Get $\mathbf{z}_{t_{i-1}}$ via solving the reverse process
\EndFor
\State $\hat{\mathbf{x}} \sim p_{\xi}\left(\mathbf{x} \mid \mathbf{z}\right)$
\State \Return $\hat{{\mathbf{S}}}$
\end{algorithmic}
\label{ag:sampling}
\end{algorithm}

%%%%%%%%%%%%%%%%%%%%%%%
\subsection{Latent Diffusion Model}
\label{sec:ldm}
After the preparations mentioned above, a trained VAE allows us to access the latent space $\mathbf{L}=\{\mathbf{z}\}_{i=1}^N$. 
% Consequently, we are ready to design a diffusion model to generate the latent space information.
Figure~\ref{fig:timeLDM}(c) presents the neural network architecture of LDM.
First, we reshape the sampling representation into one dimension before passing through a linear layer. Next, we transform the time step $t$ into sinusoidal embeddings $\boldsymbol{t}_{\text{emb}}$, and added to the $\text{Linear}(\mathbf{z})$ . After that, we apply four linear layers to learn the denoising pattern.
Finally, we reshape the latent representation back to the input shape.
Following~\cite{song2020score}, we adopt below  forward process Equation~\ref{eq:forward} and reverse process Equation~\ref{eq:reverse} to obtain noising data and learn to reverse back:
\begin{equation}
\mathbf{z}_t=\mathbf{z}_0+\sigma(t) \cdot {\varepsilon},  \varepsilon \sim \mathcal{N}(\mathbf{0}, \mathbf{I})
\label{eq:forward}
\end{equation}
\begin{equation}
\mathrm{d} \mathbf{z}_t=-2 \dot{{\sigma}}(t) \sigma(t) \nabla_{\mathbf{z}_t} \log p\left(\mathbf{z}_t\right) \mathrm{d} t+\sqrt{2 \dot{\sigma}(t) \sigma(t)} \mathrm{d} \mathbf{\omega}_t
\label{eq:reverse}
\end{equation}
where $\mathbf{z}_0=\mathbf{z}$ is the original latent representation from encoder, $\mathbf{z}_t$ is diffused representation with noise level $\sigma(t)$. 
While for reverse process, $\nabla_{\mathbf{z}_t} \log p_t\left(\mathbf{z}_t\right)$ preset score of the $\mathbf{z}_t$, $\omega_t$ is the standard Wiener process. 
The training object of LDM is:
\begin{equation}
\left.\mathcal{L}_{\text {LDM}}=\mathbb{E}_{\mathbf{z}_0 \sim p\left(\mathbf{z}_0\right)} \mathbb{E}_{t \sim p(t)} \mathbb{E}_{{\varepsilon} \sim \mathcal{N}(\mathbf{0}, \mathbf{I})} \| \mathbf{\epsilon}_\theta\left(\mathbf{z}_t, t\right)-\mathbf{\varepsilon}\right \|_2^2
\label{eq:ldm loss}
\end{equation}
where  $\boldsymbol{\epsilon}_\theta$ is the neural network to project $\mathbf{z}_t$ into Gaussian noise.
Following~\cite{karras2022elucidating}, we set the noise level $\sigma(t)=t$, and $\nabla_{\mathbf{z}_t} \log p\left(\mathbf{z}_t\right)=-\mathbf{\epsilon}_\theta\left(\mathbf{z}_t, t\right) / \sigma(t)$.

\begin{table}[b!]
\centering
\resizebox{\columnwidth}{!}{%
\begin{tabular}{ccccccccc}
\hline Parameter & Sines & MuJoCo & Stocks &  ETTh  & fMRI \\
\hline 
dim(x) & 5 & 14 & 6 & 7 & 50 \\
Attention heads & 2 & 2 & 2 & 2   & 2 \\
Attention head dimension & 16 & 16 & 16 & 16  & 16\\
Encoder layers & 1 & 1 & 2 & 2  & 1 \\
Decoder layers & 2 & 2 & 3 & 3 & 2 \\
Batch size & 1024 & 1024 & 512 & 1024  & 1024 \\
Hidden dimension of LDM & 1024 & 4096 & 1024 & 1024 & 4096 \\
\hline
\end{tabular}}
\caption{Hyperparameters of VAE and LDM. }
\label{tab:unconditional-HP}
\end{table}
%%%%%%%%%%%%%%%%%%%%%%%
\subsection{Training and Sampling}
\label{sec:TS}
With the proposed formulation and practical parameterization, we now introduce the training and sampling schemes for TimeLDM. 
The training process of TimeLDM can be divided into two steps where the first step is to train the VAE and the second step is to study the LDM on the latent space. 
The Algorithm~\ref{ag:training} presents the overall training procedure. 
For sampling process includes generative diffusion data on the standard latent states, reversing the original time series with a well-learning decoder. The Algorithm~\ref{ag:sampling} shows the overall sampling process.

%%%%%%%%%%%%%%%%%%%%%%
\section{EXPERIMENTS}

We evaluate TimeLDM for time series generation with five different benchmarks, covering simulated and real-world datasets. Our framework demonstrates better performance than existing methods, both qualitatively and quantitatively. Further analysis across various lengths of time series data confirms the robustness of TimeLDM.

\begin{table}[t]
\centering
\resizebox{0.8\columnwidth}{!}{%
\begin{tabular}{c|c|c|c}
\hline Metric       & Methods        & Sines                                                        & MuJoCo                           \\
\hline 
\rowcolor{gray!30}
                   & TimeLDM        & \textbf{0.004$\pm$.001}                     & \textbf{0.006$\pm$.000}       \\
                   & Diffusion-TS   & \underline{0.006$\pm$.000}               & \underline{0.013$\pm$.001}    \\
Context-FID        & TimeGAN        & 0.101$\pm$.014                                      & 0.563$\pm$.052               \\
Score              & TimeVAE        & 0.307$\pm$.060                                       & 0.251$\pm$.015             \\
                   % & Diffwave       & 0.014$\pm$.002                                       & 0.393$\pm$.041               \\
                   & DiffTime       & 0.006$\pm$.001                                     & 0.188$\pm$.028              \\
(Lower the Better) & Cot-GAN        & 1.337$\pm$.068                                      & 1.094$\pm$.079                \\
\hline 
\rowcolor{gray!30}
                   & TimeLDM        & \textbf{0.013$\pm$.005}                     & \textbf{0.189$\pm$.029}     \\
                   & Diffusion-TS   & \underline{0.015$\pm$.004}              & \underline{0.193$\pm$.027}   \\
Correlational      & TimeGAN        & 0.045$\pm$.010                                     & 0.886$\pm$.039                \\
Score              & TimeVAE        & 0.131$\pm$.010                                     & 0.388$\pm$.041               \\
                   % & Diffwave       & 0.022$\pm$.005                                       & 0.579$\pm$.018                \\
                   & DiffTime       & 0.017$\pm$.004                                    & 0.218$\pm$.031                \\
(Lower the Better) & Cot-GAN        & 0.049$\pm$.010                                    & 1.042$\pm$.007              \\
\hline 
\rowcolor{gray!30}
                   & TimeLDM        & \textbf{0.006$\pm$.005}                   & \textbf{0.004$\pm$.004}       \\
                   & Diffusion-TS   & \underline{0.006$\pm$.007}               & \underline{0.008$\pm$.002}   \\
Discriminative     & TimeGAN        & 0.011$\pm$.008                                  & 0.238$\pm$.068                \\
Score              & TimeVAE        & 0.041$\pm$.044                                      & 0.230$\pm$.102                \\
                   % & Diffwave       & 0.017$\pm$.008                                    & 0.203$\pm$.096                \\
                   & DiffTime       & 0.013$\pm$.006                                     & 0.154$\pm$.045                \\
(Lower the Better) & Cot-GAN        & 0.254$\pm$.137                                       & 0.426$\pm$.022               \\
\hline
\rowcolor{gray!30}
                   & TimeLDM        & \textbf{0.093$\pm$.000}                     & \textbf{0.007$\pm$.000}    \\
                   & Diffusion-TS   & \underline{0.093$\pm$.000}              & \underline{0.007$\pm$.000}      \\
                   & TimeGAN        & 0.093$\pm$.019                                     & 0.025$\pm$.003                \\
Predictive         & TimeVAE        & \underline{0.093$\pm$.000}                        & 0.012$\pm$.002              \\
% Score              & Diffwave       & \underline{0.093$\pm$.000}                        & 0.013$\pm$.000              \\
Score                  & DiffTime       & \underline{0.093$\pm$.000}                          & 0.010$\pm$.001               \\
(Lower the Better) & Cot-GAN        & 0.100$\pm$.000  & 0.068$\pm$.009\\
\cmidrule(lr){2-4}
                & Original & 0.094$\pm$.001 & 0.007$\pm$.001  \\
\hline
\end{tabular}}
\caption{Main results on simulated time series datasets. The best result in each case is \textbf{bolded}.}
% % \vspace{10pt}
\label{tab:unconditional-TS1}
\end{table}

\begin{figure}[htbp]
\centering
\resizebox{0.8\columnwidth}{!}{%
\includegraphics{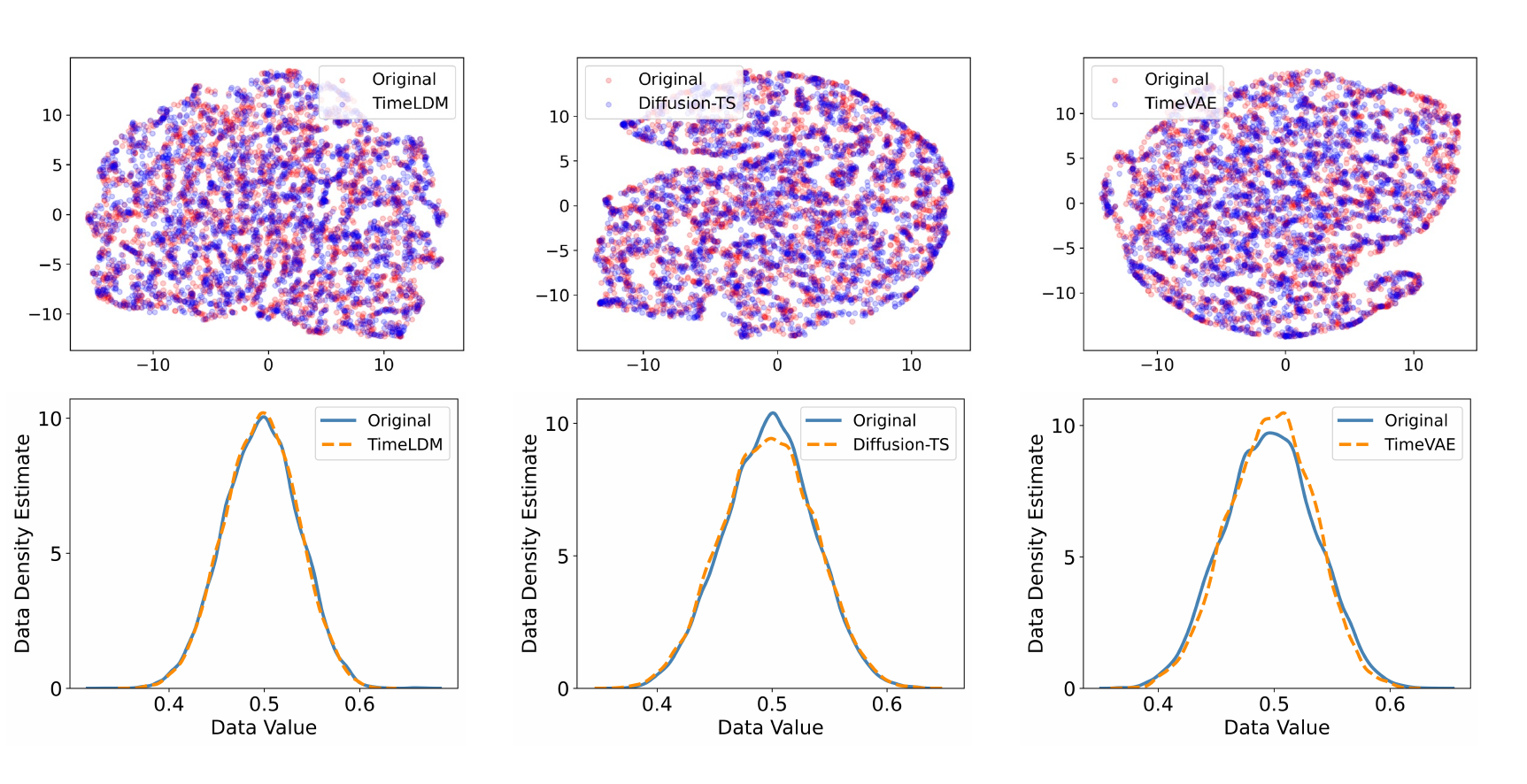}}
\caption{Visualizations of the simulated MuJoCo dataset, synthesized by TimeLDM, Diffusion-TS and TimeVAE.} 
\label{fig:kernel2}
\end{figure}

%%%%%%%%%%%%%%%%%%%%%%%
\subsection{Experimental Setups}
\label{sec:es}
%%%%%%%%%%%%%%%%%%%%%%%
% \vspace{10pt}
\noindent\textbf{Datasets.}
We utilize five different datasets to evaluate our model, including:
{\noindent\textit{Sine}} is a simulated dataset with 5 features in sinusoidal sequence, and each feature has independent frequencies and phases~\cite{yoon2019time};
{\noindent\textit{MuJoCo}} is the multivariate physics simulation time series data with 14 features~\cite{tunyasuvunakool2020dm_control};
{\noindent\textit{Stocks}} is the Google stock price information from 2004 to 2019, presented daily information and includes 6 features~\cite{desai2021timevae};
{\noindent\textit{ETTh}} is built from electricity transformers on 15 minutes basis, including load and oil temperature from July 2016 to July 2018~\cite{zhou2021informer};
{\noindent\textit{fMRI}}
 serves as a benchmark for causal discovery, featuring simulations that realistically mimic blood-oxygen-level-dependent time series~\cite{smith2011network}. 
 % In this instance, we have selected a simulation from the original dataset that comprises 50 features.

%%%%%%%%%%%%%%%%%%%%%%%
\noindent\textbf{Baseline.}
We compare our TimeLDM against five unconditional time series generation methods, including \textit{Diffusion-based} architectures (Diffusion-TS~\cite{yuan2024diffusion} and DiffTime~\cite{coletta2024constrained}), \textit{GAN-based} models (TimeGAN~\cite{yoon2019time} and Cot-GAN~\cite{xu2020cot}), and \textit{VAE-based} approach (TimeVAE~\cite{desai2021timevae}).

%%%%%%%%%%%%%%%%%%%%%%%%%%%%%%%%%%%%%%%%%%%%%%%%%%%%%%%%%%%

%%%%%%%%%%%%%%%%%%%%%%%%%%%%%%%%%%%%%%%%%%%%%%%%%%%%%%%%%%%
\begin{table}[t]
\centering
\resizebox{\columnwidth}{!}{%
\begin{tabular}{c|c|c|c|c}
\hline Metric       & Methods                                   & Stocks                            & ETTh                            & fMRI \\
\hline 
\rowcolor{gray!30}
                   & TimeLDM                  & \textbf{0.032$\pm$.007}          & \textbf{0.034$\pm$.003}               & \underline{0.139$\pm$.025} \\
                   & Diffusion-TS           & 0.147$\pm$.025                   & \underline{0.116$\pm$.010}           & \textbf{0.105$\pm$.006} \\
Context-FID        & TimeGAN                          & \underline{0.103$\pm$.013}       & 0.300$\pm$.013                               & 1.292$\pm$.218\\
Score              & TimeVAE                            & 0.215$\pm$.035                   & 0.805$\pm$.186                               & 14.449$\pm$.969\\
                   % & Diffwave                          & 0.232$\pm$.032                   & 0.873$\pm$.061                              & 0.244$\pm$.018 \\
                   & DiffTime                          & 0.236$\pm$.074                   & 0.299$\pm$.044                                & 0.340$\pm$.015\\
(Lower the Better) & Cot-GAN                           & 0.408$\pm$.086                   & 0.980$\pm$.071                               & 7.813$\pm$.550\\
\hline 
\rowcolor{gray!30}
                   & TimeLDM                 & {0.028$\pm$.009}                    & \textbf{0.028$\pm$.009}                & \textbf{1.036$\pm$.025}\\
                   & Diffusion-TS          & \textbf{0.004$\pm$.001}           & \underline{0.049$\pm$.008}           & \underline{1.411$\pm$.042}\\
Correlational      & TimeGAN                           & 0.063$\pm$.005                    & 0.210$\pm$.006                                   & 23.502$\pm$.039 \\
Score              & TimeVAE                           & 0.095$\pm$.008                    & 0.111$\pm$.020                                  & 17.296$\pm$.526\\
                   % & Diffwave                        & 0.030$\pm$.020                    & 0.175$\pm$.006                               & 3.927$\pm$.049 \\
                   & DiffTime                       & \underline{0.006$\pm$.002}        & 0.067$\pm$.005                             & 1.501$\pm$.048 \\
(Lower the Better) & Cot-GAN                          & 0.087$\pm$.004                    & 0.249$\pm$.009                                   & 26.824$\pm$.449\\
\hline 
\rowcolor{gray!30}
                   & TimeLDM                  & \textbf{0.017$\pm$.011}           & \textbf{0.009$\pm$.003}                & \textbf{0.102$\pm$.020}\\
                   & Diffusion-TS          & \underline{0.067$\pm$.015}        & \underline{0.061$\pm$.009}          & \underline{0.167$\pm$.023}\\
Discriminative     & TimeGAN                          & 0.102$\pm$.021                    & 0.114$\pm$.055                             & 0.484$\pm$.042 \\
Score              & TimeVAE                          & 0.145$\pm$.120                    & 0.209$\pm$.058                                & 0.476$\pm$.044 \\
                   % & Diffwave                          & 0.232$\pm$.061                    & 0.190$\pm$.008                             & 0.402$\pm$.029\\
                   & DiffTime                         & 0.097$\pm$.016                    & 0.100$\pm$.007                             & 0.245$\pm$.051\\
(Lower the Better) & Cot-GAN                          & 0.230$\pm$.016                    & 0.325$\pm$.099                               & 0.492$\pm$.018 \\
\hline
\rowcolor{gray!30}
                   & TimeLDM                 & \underline{0.037$\pm$.000}        & \textbf{0.118$\pm$.007}               & \textbf{0.099$\pm$.000}\\
                   & Diffusion-TS           & \textbf{0.036$\pm$.000}           & \underline{0.119$\pm$.002}            & \underline{0.099$\pm$.000}\\
                   & TimeGAN                          & 0.038$\pm$.001                    & 0.124$\pm$.001                                 & 0.126$\pm$.002\\
Predictive         & TimeVAE              & 0.039$\pm$.000                    & 0.126$\pm$.004                                & 0.113$\pm$.003 \\
% Score              & Diffwave             & 0.047$\pm$.000                    & 0.130$\pm$.001                                 & 0.101$\pm$.000\\
Score                 & DiffTime             & 0.038$\pm$.001                    & 0.121$\pm$.004                                  & 0.100$\pm$.000\\
(Lower the Better) & Cot-GAN       & 0.047$\pm$.001 & 0.129$\pm$.000 & 0.185$\pm$.003\\
\cmidrule(lr){2-5}
                & Original & 0.036$\pm$.001 & 0.121$\pm$.005   & 0.090$\pm$.001 \\
\hline
\end{tabular}}
\caption{Main results on real-world time series datasets. The best result in each case is \textbf{bolded}.}
% % \vspace{10pt}
\label{tab:unconditional-TS2}
\end{table}

%%%%%%%%%%%%%%%%%%%%%%%
\noindent\textbf{Setups.}
In this paper, all the neural networks of TimeLDM are implemented with PyTorch~\cite{paszke2019pytorch} package. For the well-training across all datasets, we tune the limited hyperparameter, as shown in Table~\ref{tab:unconditional-HP}. We proceed with the training in two steps. The first step is to train the $\beta$-VAE to obtain latent space information. The second step involves training a diffusion model in the latent space. 
In the first step, we optimize our network using Adam with default decay rates. The initial learning rate is $10^{-3}$. 
In the second step, We optimize our network using the Adam optimizer with the first and second moment decay rates set to \(0.9\) and \(0.96\) respectively. The initial learning rate start is $10^{-4}$. 
The main results are training on an NVIDIA RTX 4080 GPU.

\begin{figure}[t]
\centering
\resizebox{0.8\columnwidth}{!}{%
\includegraphics{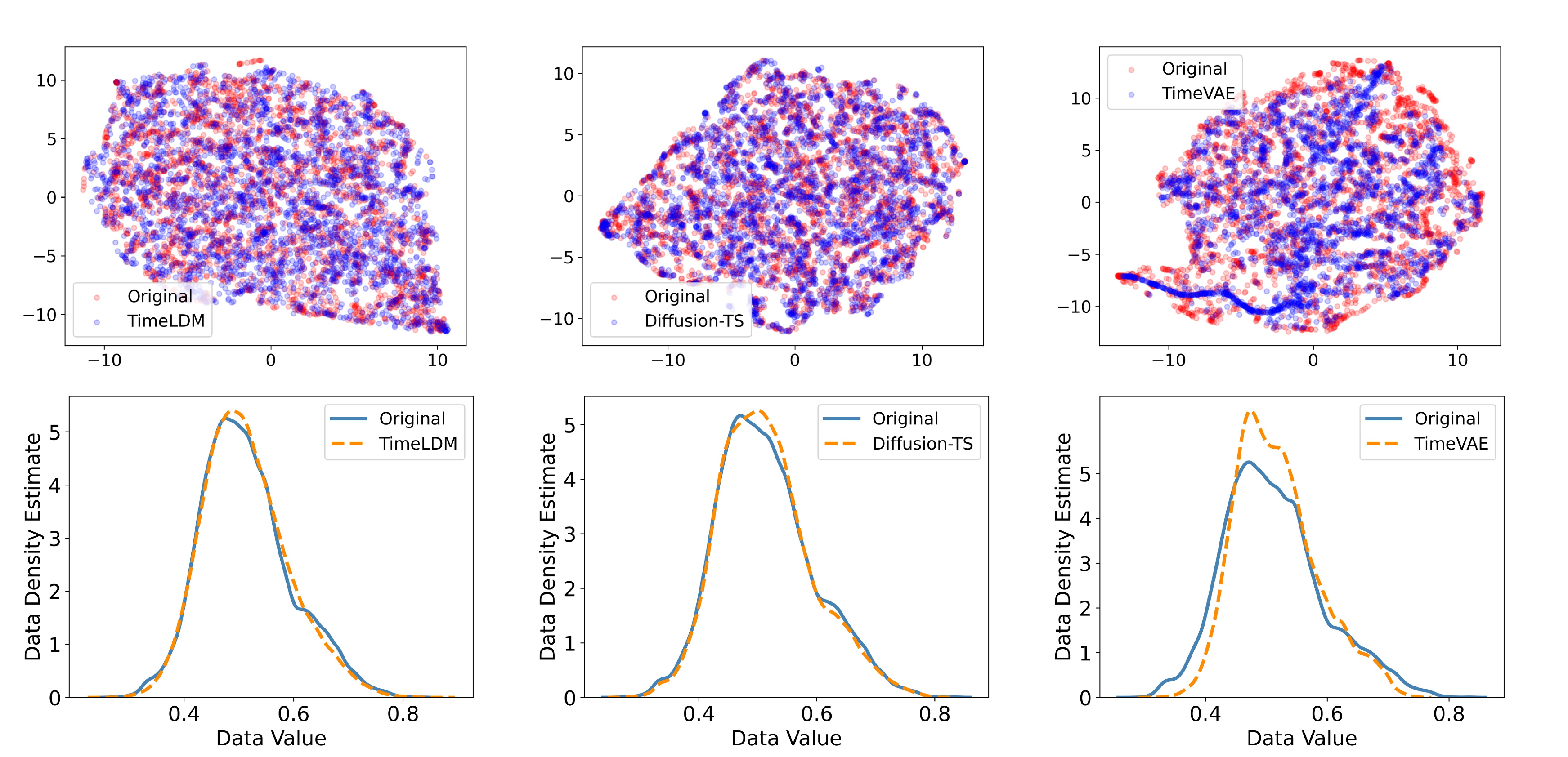}}
\caption{Visualizations of the real-world ETTh dataset, synthesized by TimeLDM, Diffusion-TS, and TimeVAE.} 
\label{fig:kernel3}
\end{figure}

%%%%%%%%%%%%%%%%%%%%%%%
\noindent\textbf{Evaluation Methods.}
For quantitative analysis,
we adopt four different evaluation metrics to evaluate the synthesized time series:
(1) \textit{Context-Fréchet Inception Distance (Context-FID) score} assesses the quality of the synthetic time series samples by calculating  the difference between representations of time series that fit into the local context~\cite{jeha2021psa};
(2) \textit{Correlational score} assesses temporal dependencies by calculating the absolute error between the cross-correlation matrices of real and synthetic data~\cite{liao2020conditional};
(3) \textit{Discriminative score} evaluates similarity by employing a classification model to differentiate between original and synthetic data in a supervised setting~\cite{yoon2019time};
(4) \textit{Predictive score} assesses the utility of synthesized data by training a sequence model post-hoc to predict future temporal vectors using the train-synthesis-and-test-real (TSTR) method~\cite{yoon2019time}.
For qualitative analysis,
we apply two different data representation methods to evaluate the synthesized time series:
(1) \textit{t-SNE} evaluates synthesized time series by projecting both original and synthetic data into a two-dimensional space~\cite{van2008visualizing};
(2) \textit{Kernel density estimation} is to draw data distributions to check the alignment between original and synthetic data.

% 0.138 
% 0.309 
\begin{figure*}[htpb]
\centering
\resizebox{\textwidth}{!}{%
\includegraphics{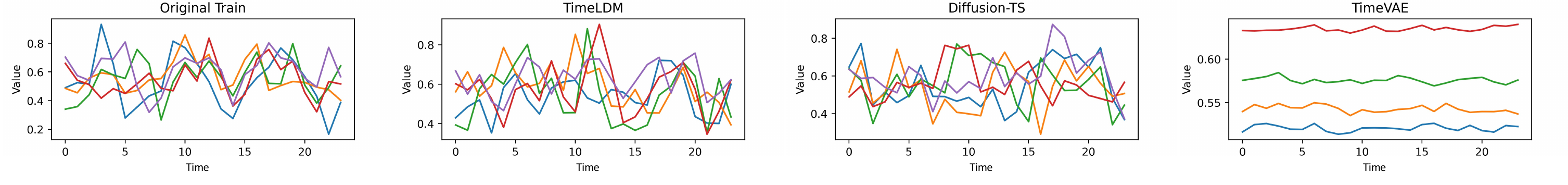}}
\caption{Examples of generating time series from the fMRI dataset. Our approach yields the closest results to the original training data.} 
\label{fig:kernel4}
\end{figure*}

%%%%%%%%%%%%%%%%%%%%%%%
\subsection{Unconditional Time Series Generation}
\label{sec:pf}
% % \vspace{10pt}
\noindent\textbf{Main Results.}
We follow the previous setup in TimeGAN~\cite{yoon2019time} to analyze the performance of models on the benchmark datasets mentioned above. 
The quantitative evaluation results of 24-length time series generation, which represents the most common comparison in existing works, are listed in Table~\ref{tab:unconditional-TS1} and~\ref{tab:unconditional-TS2}. As can be seen, TimeLDM achieves state-of-the-art results on the simulated benchmarks. Compared with the discriminative score, TimeLDM achieves an average improvement of \textbf{55\%} over Diffusion-TS~\cite{yuan2024diffusion} in all benchmarks.
% with the average improvement of 3.8$\times$ in the Discriminative score.
Figure~\ref{fig:kernel2} and~\ref{fig:kernel3} show the qualitative evaluation results of t-SNE and Kernel density. For the t-SNE analysis, 
where a greater overlap of blue and red dots shows a better distributional similarity between the generated data and original data.
Figure~\ref{fig:kernel2} and~\ref{fig:kernel3} of t-SNE reveals that our methods have better overlap between the generated data and original data.
The Kernel density presents the distribution alignment of original data and synthetic information. Based on the figure, our TimeLDM aligns better with the original data than Diffusion-TS and TimeVAE. We also present the generating time series from the fMRI dataset in Figure~\ref{fig:kernel4}. Compared to the Diffusion-TS~\cite{yuan2024diffusion} and TimeVAE~\cite{desai2021timevae}, TimeLDM generates time series that more closely resemble the original training set, while TimeVAE~\cite{desai2021timevae} struggles to learn features from the fMRI dataset.

\begin{table}[b!]
\centering
\resizebox{0.8\columnwidth}{!}{%
\begin{tabular}{c|c|c|c}
\hline Metric & Methods & ETTh-64 & ETTh-128   \\
\hline 
\rowcolor{gray!30}
& TimeLDM &  \textbf{0.067$\pm$.008} & \textbf{0.169$\pm$.015}\\
& Diffusion-TS  & \underline{0.631$\pm$.058} &\underline{0.787$\pm$.062}  \\
Context-FID & TimeGAN & 1.130$\pm$.102 & 1.553$\pm$.169  \\
Score & TimeVAE  & 0.827$\pm$.146 & 1.062$\pm$.134    \\
% & Diffwave  & 1.543$\pm$.153 & 2.354$\pm$.170  \\
& DiffTime  &  1.279$\pm$.083 & 2.554$\pm$.318    \\
(Lower the Better) & Cot-GAN  &3.008$\pm$.277 & 2.639$\pm$.427    \\

\hline 
\rowcolor{gray!30}
& TimeLDM & \textbf{0.034$\pm$.005} &  \underline{0.058$\pm$.010}  \\
& Diffusion-TS & \underline{0.082$\pm$.005} & {0.088$\pm$.005}  \\
Correlational & TimeGAN & 0.483$\pm$.019 & 0.188$\pm$.006   \\
Score& TimeVAE &  0.067$\pm$.006 & \textbf{0.054$\pm$.007} \\
% & Diffwave &  0.186$\pm$.008 &  0.203$\pm$.006 \\
& DiffTime & 0.094$\pm$.010 &0.113$\pm$.012  \\
(Lower the Better) & Cot-GAN & 0.271$\pm$.007 & 0.176$\pm$.006   \\
\hline 
\rowcolor{gray!30}
& TimeLDM &  \textbf{0.030$\pm$.053}  & \textbf{0.080$\pm$.044}   \\
& Diffusion-TS &  \underline{0.106$\pm$.048}&\underline{0.144$\pm$.060}  \\
Discriminative & TimeGAN & 0.227$\pm$.078&  0.188$\pm$.074   \\
Score & TimeVAE & 0.171$\pm$.142 &0.154$\pm$.087   \\
% & Diffwave &  0.254$\pm$.074  & 0.274$\pm$.047   \\
& DiffTime & 0.150$\pm$.003  & 0.176$\pm$.015   \\
(Lower the Better) & Cot-GAN & 0.296$\pm$.348 &0.451$\pm$.080   \\

\hline
\rowcolor{gray!30}
& TimeLDM &\textbf{0.115$\pm$.010} &  \underline{0.117$\pm$.009}   \\
& Diffusion-TS & \underline{0.116$\pm$.000} & \textbf{0.110$\pm$.003}   \\
& TimeGAN &  0.132$\pm$.008 & 0.153$\pm$.014  \\
Predictive & TimeVAE & 0.118$\pm$.004 & 0.113$\pm$.005 \\
% Score & Diffwave & 0.133$\pm$.008 & 0.129$\pm$.003  \\
Score& DiffTime &0.118$\pm$.004 & 0.120$\pm$.008   \\
(Lower the Better) & Cot-GAN & 0.135$\pm$.003 & 0.126$\pm$.001  \\
\hline
\end{tabular}}
\caption{Further evaluation on long-term ETTh dataset generation. The best result in each case is \textbf{bolded}.}
\label{tab:lunconditional-TS1}
\end{table}

\begin{figure}[h]
\centering
\resizebox{\columnwidth}{!}{%
\includegraphics{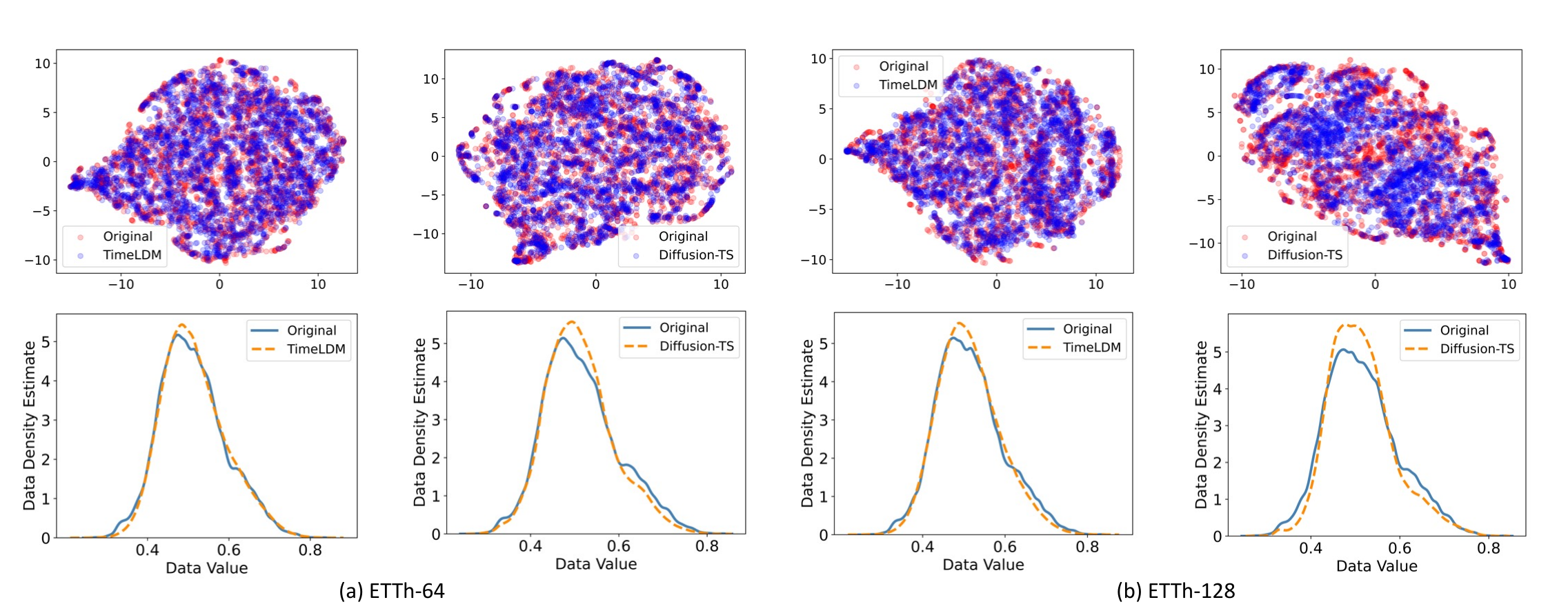}}
\caption{Visualizations of the real-world ETTh dataset, synthesized by TimeLDM and Diffusion-TS with 64 and 128-time series lengths.}
\label{fig: ltsne5}
% % \vspace{10pt}
\end{figure}

\noindent\textbf{Further Analysis.}
To further confirm the scalability of our TimeLDM, we evaluate the impact of the different time-series lengths on the generative models for unconditional time series.  We examine  ETTh data with two different
lengths, 64 and 128. For these experiments, we keep all the same hyperparameters with the same metrics to assess the generation quality of different methods. The quantitative results are reported in Table~\ref{tab:lunconditional-TS1}.
As we can see, our proposed TimeLDM can achieve better performance in most evaluation metrics. Especially on the Context-FID score and Discriminative score, TimeLDM realizes significant state-of-the-art performance with \textbf{80\%} and \textbf{50\%} improvement over Diffusion-TS~\cite{yuan2024diffusion}.
The qualitative results are depicted in Figure~\ref{fig:kernel4}. Our TimeLDM shows better alignment than Diffusion-TS with the original data. 

%0.236 0.11
% 1.418  0.25

%%%%%%%%%%%%%%%%%%%%%%%
\subsection{Ablation Study}
\label{sec:as}
In this part,  we first assess the adaptive $\beta$ with the fixed values ($\beta_{\max }$, $\beta_{\min }$) in the VAE model. Then, we analyze the effectiveness of the reconstruction loss function of VAE. We compare the loss function with its three variants: (1) w/o FFT loss term during training, (2) w/o $\mathcal{L}_1$ norm term during training, (3) w/o $\mathcal{L}_2$ norm term during training. The ablation study across all the benchmarks presents the results in Table~\ref{tab:unconditional-TS-ab1} and Table~\ref{tab:unconditional-TS-ab2}, respectively. 

% % \vspace{10pt}
\noindent\textbf{The effect of adaptive $\beta$.}
We evaluate the adaptive weighting coefficient $\beta$ in the VAE model. Table~\ref{tab:unconditional-TS-ab1} presents the results of adaptive $\beta$ and constant values ($\beta_{\max }$, $\beta_{\min }$) on the aforementioned datasets.
As can be seen, the difference in performance between the Sines and Stocks benchmarks is insignificant. At the same time, there is a significant performance disparity between the ETTh and MuJoCo benchmarks. 
The adaptive $\beta$ improve the effectiveness of TimeLDM, remarkably. This emphasizes the superior performance demonstrated by the adaptive $\beta$  approach in training the VAE model.

\begin{table}
\centering
\resizebox{\columnwidth}{!}{%
\begin{tabular}{c|c|c|c|c|c|c}
\hline Metric & $\beta$ & Sines & MuJoCo & Stocks & ETTh  & fMRI  \\
\hline 
\rowcolor{gray!30}
Discriminative & Adaptive & {0.006$\pm$.005} & \textbf{0.004$\pm$.004} & \underline{0.017$\pm$.011} & \textbf{0.009$\pm$.003}   & \textbf{0.102$\pm$.020} \\
 Score& $10^{-2}$& \textbf{0.004$\pm$.004} & 0.258$\pm$.023 & \textbf{0.015$\pm$.015} & {0.034$\pm$.019}  & 0.496$\pm$.003 \\
(Lower the Better) &  $10^{-5}$  & \textbf{0.004$\pm$.004} & {0.007$\pm$.008} & {0.020$\pm$.019} & 0.093$\pm$.004 &  0.216$\pm$.021\\
\hline
\rowcolor{gray!30}
& Adaptive & \textbf{0.093$\pm$.000} & \textbf{0.007$\pm$.000} & \textbf{0.037$\pm$.000} &  \textbf{0.118$\pm$.007}  &  \textbf{0.099$\pm$.000}\\
Predictive & $10^{-2}$ & \textbf{0.093$\pm$.000} & 0.013$\pm$.001 & \textbf{0.037$\pm$.000} & {0.122$\pm$.005}   &  0.100$\pm$.000\\
Score&  $10^{-5}$ & \textbf{0.093$\pm$.000} & {0.007$\pm$.001} & {0.038$\pm$.000} & 0.123$\pm$.005   & 0.100$\pm$.000 \\
\cmidrule(lr){2-7}
(Lower the Better) &Original & 0.094$\pm$.001 & 0.007$\pm$.001 & 0.036$\pm$.001 & 0.121$\pm$.005 &  0.090$\pm$.001 \\
\hline
\end{tabular}}
\caption{Ablation study for the adaptive $\beta$, which balances the reconstruction loss and KL loss. The best result in each case is \textbf{bolded}.}
% % \vspace{10pt}
\label{tab:unconditional-TS-ab1}
\end{table}

\begin{table}
\centering
\resizebox{\columnwidth}{!}{%
\begin{tabular}{c|c|c|c|c|c|c}
\hline Metric & Methods & Sines & MuJoCo & Stocks & ETTh & fMRI  \\
\hline 
\rowcolor{gray!30}
& TimeLDM & \textbf{0.006$\pm$.005} &\textbf{0.004$\pm$.004} & {0.017$\pm$.011} & \textbf{0.009$\pm$.003}   & \textbf{0.102$\pm$.020}\\
Discriminative & w/o FFT & 0.008$\pm$.003 &  {0.005$\pm$.002} & 0.022$\pm$.028 & 0.013$\pm$.009  & 0.115$\pm$.019 \\
Score & w/o $\mathcal{L}_1$ & {0.007$\pm$.005} & {0.008$\pm$.004} & 0.020$\pm$.015 & 0.014$\pm$.014  & 0.107$\pm$.019\\
(Lower the Better) &  w/o $\mathcal{L}_2$  & \textbf{0.006$\pm$.005} & 0.007$\pm$.006 & \textbf{0.014$\pm$.010} & {0.010$\pm$.008}  & 0.129$\pm$.014 \\
\hline
\rowcolor{gray!30}
& TimeLDM & \textbf{0.093$\pm$.000} & \textbf{0.007$\pm$.000} & \textbf{0.037$\pm$.000} & {0.118$\pm$.007}  &  \textbf{0.099$\pm$.000}\\
 & w/o FFT & \textbf{0.093$\pm$.000} & {0.008$\pm$.002} & \textbf{0.037$\pm$.000} & \textbf{0.118$\pm$.006}   & \textbf{0.099$\pm$.000} \\
Predictive & w/o $\mathcal{L}_1$ & \textbf{0.093$\pm$.000} & {0.008$\pm$.002} & \textbf{0.037$\pm$.000} & 0.121$\pm$.004  &  \textbf{0.099$\pm$.000} \\
Score&  w/o $\mathcal{L}_2$  & \textbf{0.093$\pm$.000} & {0.007$\pm$.001} & \textbf{0.037$\pm$.000} & 0.121$\pm$.006 &  \textbf{0.099$\pm$.000} \\
\cmidrule(lr){2-7}
(Lower the Better) &Original & 0.094$\pm$.001 & 0.007$\pm$.001 & 0.036$\pm$.001 & 0.121$\pm$.005 &  0.090$\pm$.001 \\
\hline
\end{tabular}}
\caption{Ablation study for VAE reconstruction loss function. The best result in each case is \textbf{bolded}.}
% % \vspace{10pt}
\label{tab:unconditional-TS-ab2}
\end{table}

% \vspace{10pt}
\noindent\textbf{The effect of reconstruction loss.}
We evaluate the effectiveness of reconstruction loss term in the VAE model. Table~\ref{tab:unconditional-TS-ab2} presents the results of three variants on the aforementioned datasets.
As can be seen,
the performance gap among them for the Predictive score is negligible, the FFT loss term and $\mathcal{L}_1$ norm term show a crucial role in improving the capability of TimeLDM for the Discriminative score.

\section{CONCLUSIONS}

In this paper, we propose TimeLDM, a novel latent diffusion model for unconditional time series generation. 
Particularly, we explore diffusion on the latent space where the original time series is encoded by a variational autoencoder. 
We evaluate our method on the simulated and real-world datasets and benchmark the performance against existing state-of-the-art methods. 
Experimental results demonstrate that TimeLDM persistently delivers high-quality generated data both qualitatively and quantitatively. 
Remarkably, TimeLDM achieves new state-of-the-art results on the simulated benchmarks and an average improvement of 55\% in Discriminative score with all benchmarks.
Further studies demonstrate that our method yields better performance on different lengths of time series data generation. 
To the best of our knowledge, this is the first work to
explore the potential of the latent diffusion model for unconditional time series generation.
We hope that TimeLDM can serve as a robust baseline for generating informative time series tokens for agents learning in the field of physical AI.

\bibliographystyle{IEEEtran}
% \bibliography{IEEEexample}

\end{document}